\documentclass{article} 
\usepackage{iclr2027_conference,times}

\usepackage{amsmath,amsfonts,bm}

\def\eqref#1{equation~\ref{#1}}

\def\1{\bm{1}}

\DeclareMathAlphabet{\mathsfit}{\encodingdefault}{\sfdefault}{m}{sl}
\SetMathAlphabet{\mathsfit}{bold}{\encodingdefault}{\sfdefault}{bx}{n}

\usepackage{hyperref}
\usepackage{url}

\usepackage{graphicx}
\usepackage{amsmath}
\usepackage{multirow}
\usepackage{amssymb}
\usepackage{natbib}
\usepackage{booktabs}

\title{AIMS: Anchor-Integrated Multi-View Synthesis for Scalable Novel View Rendering}

\author{
JooHyun Park \\
Korea University \\
\texttt{james0223@korea.ac.kr} \\
\And
Hanyoung Jang \\
NCSOFT \\
\And
HyeongYeop Kang \\
Korea University \\
\texttt{siamiz\_hkang@korea.ac.kr}
}

\iclrfinalcopy 
\begin{document}

\maketitle

\lhead{}  

\begin{abstract}
Feed-forward novel view synthesis methods achieve strong generalization from posed multi-view inputs, but scaling them to large input view sets remains challenging. Transformer-based approaches that jointly process all input-view tokens incur rapidly increasing computation and memory as the number of views grows, while simple view subsampling discards potentially useful observations. We introduce \textbf{Anchor-Integrated Multi-View Synthesis (AIMS)}, a scalable framework that decouples the number of available observations from the number of views processed by the global synthesis model. AIMS selects a fixed set of spatially distributed anchor views using farthest point sampling, groups nearby observations around each anchor, and uses a lightweight learnable integrator to fuse their information into enriched anchor representations. This allows additional observations to contribute to synthesis while keeping the downstream global view budget fixed. Evaluations on RealEstate10K and ScanNet demonstrate a favorable quality--efficiency trade-off against transformer-based and Gaussian-based baselines. AIMS achieves 29.41\,dB and 17.73\,dB PSNR on the two datasets, respectively, with rendering averaging 7.24\,ms per view.
\end{abstract}

\section{Introduction}
\label{sec:introduction}
All views are useful, but not all views need attention. Additional observations can reveal occluded surfaces, reduce geometric ambiguity, and provide richer evidence for novel view synthesis. Yet in transformer-based models, allowing every observation to participate directly in global attention makes computation grow with the input collection. We ask whether a model can benefit from many observations while reasoning globally over only a compact representation of them.

Recent progress in feed-forward novel view synthesis has made this question increasingly relevant. Many approaches rely on explicit 3D representations or geometry-aware priors, such as neural radiance fields, 3D Gaussians, and epipolar constraints. LVSM~\citep{jin2025lvsm} takes a different direction, formulating novel view synthesis as a largely data-driven transformer problem with minimal 3D inductive bias. Given posed input images, it directly learns to synthesize target views without relying on a predefined 3D representation or rendering equation.

This flexibility, however, makes the size of the visual context an important computational bottleneck. In the decoder-only LVSM, input and target tokens interact through full self-attention, causing computation to grow rapidly as additional views are introduced. Efficient-LVSM~\citep{jia2026efficientlvsm} substantially improves this scaling through decoupled attention, but incorporating more observations still requires processing a larger collection of input tokens. The encoder--decoder variant of LVSM may appear to avoid this issue by compressing the inputs into a fixed number of latent tokens; however, the encoder must still apply full self-attention over the input tokens before producing this fixed-size representation. As a result, its computational cost also grows with the number of input views. Thus, existing formulations still couple the amount of available scene evidence with the computation required to exploit it.

In this work, we introduce \textbf{Anchor-Integrated Multi-View Synthesis} (\textbf{AIMS}), a framework that incorporates evidence from large view collections under a \textit{fixed global attention budget}. Inspired by the Point Transformer~\citep{zhao2021point}, which handles large point sets by selecting representative points and aggregating information from their local neighborhoods, AIMS applies a similar principle to large collections of input views. Specifically, AIMS selects a compact set of representative anchor views and enriches their representations with information from neighboring observations. The resulting integrated anchors summarize a larger input collection and serve as the inputs to the global synthesis model. With a fixed anchor count and per-anchor token budget, the model's per-target global attention computation remains independent of the number of input observations. The anchor and neighbor counts provide complementary controls over scene detail and computational cost: more anchors expand the capacity of the global representation, while more neighbors supply additional evidence for local refinement. Increasing these counts can support finer scene detail at higher computational cost, whereas smaller settings favor a more compact or coarser representation with lower processing cost.

Our evaluation of AIMS on RealEstate10K~\citep{zhou2018stereo} and ScanNet~\citep{dai2017scannet} demonstrates improvements in both synthesis quality and computational efficiency. 
With 12 anchors and six views per group, AIMS achieves 29.41\,dB and 17.73\,dB PSNR, respectively, outperforming the evaluated baselines. 
Its reusable scene representation further enables efficient synthesis of additional views, with rendering averaging 7.24\,ms per view.

Our contributions are summarized as follows:
\begin{itemize}
    \item We formulate large-input novel view synthesis under a \emph{fixed global attention budget}, allowing $N$ observations to contribute through $G \ll N$ integrated anchor views while keeping the number of views participating in global synthesis fixed.

    \item We introduce \textbf{AIMS}, a hierarchical view integration framework that combines coverage-aware anchor selection with learnable local aggregation, enabling non-anchor observations to contribute without directly expanding the global attention context.

    \item We conduct comprehensive evaluations on RealEstate10K and ScanNet, demonstrating that AIMS offers a favorable trade-off between reconstruction quality and computational efficiency compared with the evaluated baselines.
\end{itemize}

\section{Related Work}
\label{sec:related_work}

\paragraph{Foundations of Novel View Synthesis.}
Early novel view synthesis (NVS) generated new views from photographs using geometric proxies~\citep{debevec1996modeling} or densely sampled light fields~\citep{levoy1996light,gortler1996lumigraph}.
Later methods improved image alignment and combination through depth-guided warping~\citep{chaurasia2013depth}, appearance flow~\citep{zhou2016view}, learned blending~\citep{hedman2018deep}, and depth prediction~\citep{choi2019extreme}.
Neural scene representations offered another route: NeRF~\citep{mildenhall2021nerf} optimizes continuous density and radiance fields through differentiable volume rendering.
Subsequent work improved anti-aliasing and view-dependent appearance~\citep{barron2021mip,verbin2024ref,barron2023zip}, while voxel grids~\citep{sun2022direct,fridovich2022plenoxels}, tensor factorization~\citep{chen2022tensorf}, and hash encoding~\citep{muller2022instant} increased efficiency.
Point-based alternatives associate neural features with 3D points in Point-NeRF~\citep{xu2022point} or rasterize anisotropic Gaussians in 3D Gaussian Splatting~\citep{kerbl3Dgaussians}, with extensions addressing aliasing~\citep{yu2024mip}, distorted cameras, and secondary rays~\citep{wu20253dgut}.
While these developments advance scene reconstruction, the cost of per-scene optimization motivates models that infer scene content from learned priors.

\paragraph{Generalizable Reconstruction and Rendering.}
Generalizable NVS learns priors across scenes to predict novel views or scene representations in a feed-forward manner.
Early examples include pixelNeRF~\citep{yu2021pixelnerf}, which conditions radiance fields on pixel-aligned features, and IBRNet~\citep{wang2021ibrnet}, which aggregates multi-view evidence.
Geometric reasoning was further developed through plane-sweep cost volumes in MVSNeRF~\citep{chen2021mvsnerf} and geometry and visibility cues in GeoNeRF~\citep{johari2022geonerf} and NeuRay~\citep{liu2022neural}.
Following advances in Gaussian rendering, feed-forward models began predicting Gaussian primitives directly: pixelSplat~\citep{charatan2024pixelsplat} uses image pairs, MVSplat~\citep{chen2024mvsplat} uses multi-view cost volumes, and DepthSplat~\citep{xu2025depthsplat} combines matching with pretrained monocular depth features.
NoPoSplat~\citep{ye2025no} extends this direction to canonical reconstruction from sparse unposed images.
LRM~\citep{hong2024lrm} and GS-LRM~\citep{zhang2024gs} scale image-to-scene prediction using transformers and triplane or Gaussian representations.
To improve multi-view processing, LaRa~\citep{chen2024lara} uses group attention over Gaussian volumes.
Alongside deterministic reconstruction, generative methods address unobserved regions through camera-conditioned diffusion~\citep{watson2023diffusion,liu2023zero}, with CAT3D~\citep{gao2024cat3d} and Stable Virtual Camera~\citep{zhou2025stable} supporting multi-view generation.

\paragraph{Scalable Transformer-Based View Synthesis.}
Alongside methods that predict predefined 3D representations, learned renderers synthesize views directly from scene context and target cameras.
Light Field Networks~\citep{sitzmann2021light} predict color from ray coordinates, while Geometry-Free View Synthesis~\citep{rombach2021geometry} and ViewFormer~\citep{kulhanek2022viewformer} use transformers to synthesize views from encoded image context.
Scene Representation Transformer~\citep{sajjadi2022scene} organizes observations into latent tokens queried by target rays, whereas GPNR~\citep{suhail2022generalizable} retains geometric guidance through epipolar feature sampling.
LVSM~\citep{jin2025lvsm} advances direct synthesis at the patch level, using encoder--decoder and decoder-only architectures with minimal 3D inductive bias.
However, extending transformer-based synthesis to larger input collections makes attention cost a central concern.
Efficient-LVSM~\citep{jia2026efficientlvsm} addresses this through intra-view input attention and target self-then-cross attention, achieving linear input-view scaling and enabling input-feature caching.
Further work pursues efficient inference through reusable contexts and unidirectional decoding in SVSM~\citep{kim2026scaling}, shared reconstruction and rendering weights with key--value caching in DVSM~\citep{sun2026dvsm}, and geometry-pretrained features with lightweight rendering in LagerNVS~\citep{szymanowicz2026lagernvs}.
LaCT's NVS model~\citep{zhang2026test} instead compresses multi-view evidence into fast weights through test-time updates.
AIMS addresses the input-scaling problem through local view integration, compressing neighboring observations into fixed-size anchor representations before global synthesis.
This allows additional views to contribute without expanding the global attention context or applying full-input self-attention.

\section{Method}
\label{sec:method}

\begin{figure}[t]
    \centering
    \includegraphics[width=\textwidth]{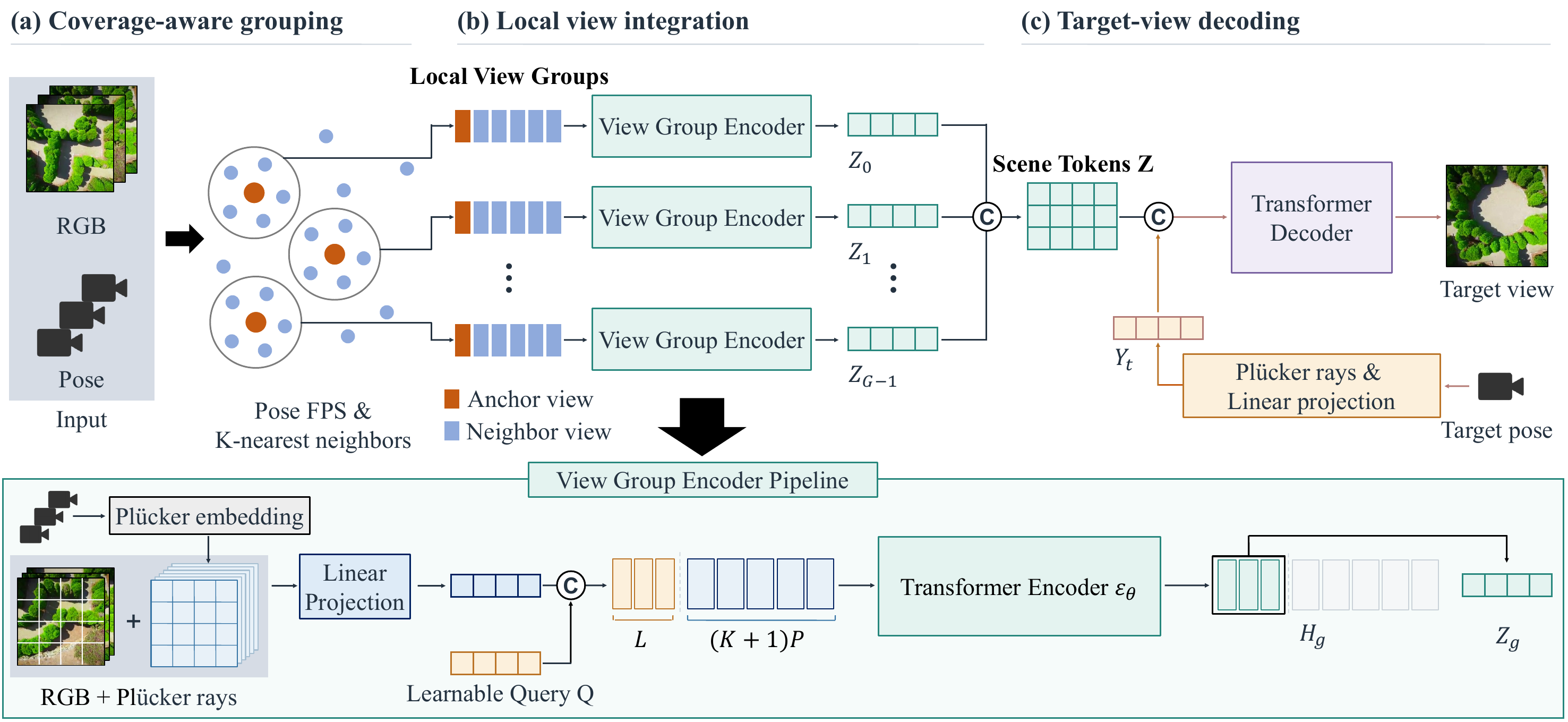} 
    \caption{Overview of AIMS. Coverage-aware grouping selects anchors and neighboring views. The View Group Encoder compresses each group's ray-conditioned image tokens into compact scene tokens. A global decoder combines these with target ray tokens to synthesize novel views.}
    \label{fig:pipeline}
\end{figure}

We introduce \textbf{Anchor-Integrated Multi-View Synthesis (AIMS)}, which synthesizes novel views from posed RGB observations through a compact scene representation. 
Given $N$ observations $\mathcal{O}=\{(I_i,\mathbf{A}_i,\mathbf{T}_i)\}_{i=1}^{N}$, where $\mathbf{A}_i$ and $\mathbf{T}_i$ denote camera intrinsics and camera-to-world poses, our goal is to predict an image $\widehat{I}_t$ at a target camera $(\mathbf{A}_t,\mathbf{T}_t)$. 
The pipeline consists of three stages:
\emph{coverage-aware view grouping}, which selects $G$ representative anchors and groups nearby observations around them;
\emph{view group encoding}, where our \textbf{View Group Encoder (VGE)} integrates ray-conditioned image tokens from each anchor and its neighbors into $L$ scene tokens; and
\emph{target-view decoding}, which synthesizes the target image from the resulting $GL$ scene tokens.
With $G$ and $L$ fixed, the global attention cost remains independent of the number of input observations $N$. 
The overall pipeline is visualized in Fig.~\ref{fig:pipeline}.

\subsection{Coverage-Aware View Grouping}
\label{sec:aims_grouping}

Inspired by the sampling and local aggregation in Point Transformer~\citep{zhao2021point}, we select representative anchor views and group nearby observations around them. Let $\mathbf{R}_i\in\mathbb{R}^{3\times3}$ and $\mathbf{c}_i\in\mathbb{R}^{3}$ denote the rotation matrix and center of camera $i$ in a normalized scene frame. Anchor selection uses deterministic farthest-point sampling in camera-pose space. We define the pose distance by combining the separation between camera centers with the angle between their orientations:
\begin{equation}
\begin{gathered}
    \delta(i,j)
    =
    \sqrt{w_{\mathrm{t}}\|\mathbf{c}_i-\mathbf{c}_j\|_2^2
          + w_{\mathrm{r}}\theta_{ij}^{2}}, \\
    \theta_{ij}
    =
    \arccos\left(
        \frac{\operatorname{tr}(\mathbf{R}_i^\top\mathbf{R}_j)-1}{2}
    \right).
\end{gathered}
    \label{eq:aims_pose_distance}
\end{equation}
The relative rotation $\mathbf{R}_i^\top\mathbf{R}_j$ satisfies $\operatorname{tr}(\mathbf{R}_i^\top\mathbf{R}_j)=1+2\cos\theta_{ij}$, where $\operatorname{tr}$ sums the diagonal entries. Solving this relation using $\arccos$ yields the smallest alignment angle $\theta_{ij}\in[0,\pi]$, a scalar measure of orientation difference in radians. The weights $w_{\mathrm{t}}$ and $w_{\mathrm{r}}$ balance this angular distance with the distance between camera centers. We use $(w_{\mathrm{t}},w_{\mathrm{r}})=(1,0.25)$.

We initialize the selected set with the pose medoid $a_1=\arg\min_i\sum_j\delta_{\mathrm{a}}(i,j)$ and iteratively select the camera farthest from the current anchors:
\begin{equation}
    a_g =
    \underset{i\notin\mathcal{A}_{g-1}}{\arg\max}
    \min_{a\in\mathcal{A}_{g-1}}\delta_{\mathrm{a}}(i,a),
    \label{eq:aims_fps}
\end{equation}
where $\mathcal{A}_{g-1}=\{a_1,\ldots,a_{g-1}\}$. This procedure encourages coverage of the available camera poses without requiring image features or reconstructed geometry.

After selecting the anchors, we form a local view group $\mathcal{G}_g$ around each anchor $a_g$ by including the anchor and its $K$ nearest neighbors under $\delta_{\mathrm{n}}$, allowing overlaps between groups. 

\subsection{View Group Encoder}
\label{sec:aims_integration}
\label{sec:aims_tokenization}

The VGE integrates the complementary observations within each local view group $\mathcal{G}_g$ into a fixed number of scene tokens.
We first tokenize the group's images together with their camera poses. 
The camera poses are encoded using per-pixel Pl\"ucker ray embeddings~\citep{plucker1865new}, computed using the corresponding
camera intrinsics.
For each source view, we concatenate corresponding RGB and ray embedding patches and linearly project them into tokens
$\mathbf{E}_i\in\mathbb{R}^{P\times D}$, where $P=HW/p^2$ is the number of $p\times p$ patches in an $H\times W$ image and $D$
is the token dimension.

We concatenate the view tokens $\mathbf{E}_i$ for $i\in\mathcal{G}_g$ into $\mathbf{X}_g\in\mathbb{R}^{(K+1)P\times D}$ and prepend
$L$ learnable tokens $\mathbf{Q}$ shared across groups.
The VGE is implemented as a shared transformer encoder $\mathcal{E}_{\theta}$ that jointly processes these tokens.
We retain the outputs corresponding to the learnable tokens:
\begin{equation}
    \mathbf{H}_g = \mathcal{E}_{\theta}([\mathbf{Q};\mathbf{X}_g]),
    \qquad
    \mathbf{Z}_g = \mathbf{H}_g[{:}L]
    \in\mathbb{R}^{L\times D},
    \label{eq:aims_group_output}
\end{equation}
where $[{:}L]$ selects the first $L$ output tokens corresponding to the learnable queries $\mathbf{Q}$.
The resulting $\mathbf{Z}_g$ summarizes the anchor and its neighbors in $\mathcal{G}_g$. 
Concatenating all group outputs gives the scene tokens $\mathbf{Z}=[\mathbf{Z}_1;\ldots;\mathbf{Z}_G]
\in\mathbb{R}^{GL\times D}$.

\subsection{Target View Decoding}
\label{sec:aims_decoding}

Following the target view decoding formulation of LVSM~\citep{jin2025lvsm}, we obtain target query tokens $\mathbf{Y}_t\in\mathbb{R}^{P_t\times D}$ by patchifying the target camera's Pl\"ucker ray embeddings and applying a separate linear projection, where $P_t$ is the number of target patches. A global transformer decoder $\mathcal{D}_{\phi}$ processes the concatenation of the target query tokens $\mathbf{Y}_t$ and scene tokens $\mathbf{Z}$, and we retain the target-token outputs:
\begin{equation}
    \mathbf{V}_t = \mathcal{D}_{\phi}([\mathbf{Y}_t;\mathbf{Z}]),
    \qquad
    \mathbf{F}_t = \mathbf{V}_t[{:}P_t]
    \in\mathbb{R}^{P_t\times D},
    \label{eq:aims_global_decoder}
\end{equation}
where $[{:}P_t]$ selects the first $P_t$ output tokens corresponding to the target queries. The retained tokens $\mathbf{F}_t$ are projected to RGB patches and assembled into the predicted image $\widehat{I}_t$.
Since $\mathbf{Z}$ is independent of the target camera, the scene tokens support synthesis for multiple target viewpoints.

By compressing each group into $L$ scene tokens, AIMS limits the sequence used for target-view decoding to $GL$ scene tokens and $P_t$ target query tokens. This gives a per-target attention cost of
\begin{equation}
    \mathrm{attn\_cost}
    =
    \mathcal{O}\left(
        (GL+P_t)^2
    \right).
    \label{eq:aims_global_cost}
\end{equation}
For fixed $G$ and $L$, the global attention cost remains independent of the input count $N$, maintaining a fixed global attention budget for target view decoding.

\subsection{Loss Function}
\label{sec:aims_loss}

For each target view, we supervise the predicted image $\widehat{I}_t$ against the ground-truth image $I_t$ using a combination of mean squared error and perceptual loss:
\begin{equation}
    \mathcal{L}
    = \operatorname{MSE}(\widehat{I}_t,I_t)
    + \lambda\,\operatorname{Perceptual}(\widehat{I}_t,I_t),
    \label{eq:aims_loss}
\end{equation}
where $\lambda=0.5$ and $\operatorname{Perceptual}$ denotes LPIPS with a VGG backbone~\citep{zhang2018unreasonable}. The MSE term encourages pixel-wise accuracy, while the perceptual term encourages visual similarity.

\section{Experiments}
\label{sec:experiments}

\subsection{Datasets}
\label{sec:aims_datasets}

\paragraph{RealEstate10K.}
We use RealEstate10K~\citep{zhou2018stereo}, which contains approximately 80K video clips curated from 10K YouTube videos covering both indoor and outdoor scenes.

\paragraph{ScanNet.}
ScanNet~\citep{dai2017scannet} contains RGB-D sequences of indoor environments with calibrated cameras. 
We include ScanNet as a complementary benchmark, as qualitative inspection of our sampled sequences suggested greater intra-scene camera-pose variation in ScanNet than in RealEstate10K.

\paragraph{Input and target views.}
We sample input and target views from the RealEstate10K test set and 843 ScanNet scenes. 
For each scene, we first uniformly sample 100 frames to form the observation set. 
We then reserve up to eight uniformly spaced views as targets. 
From the remaining views, we construct the input set using the coverage-aware grouping strategy described in Sec.~\ref{sec:aims_grouping}, with 12 anchors and 5 neighbors per anchor.
Using these inputs, AIMS integrates each local view group through the View Group Encoder (VGE) before global synthesis, while the baselines process the selected images using their original architectures.
Since the released baseline checkpoints were trained with few input views and may not generalize well to large input collections, we additionally evaluate them using only $2$, $4$, $8$, or $12$ representative anchor images selected via respective coverage-aware view grouping.

\subsection{Comparison with Baselines}
\label{sec:aims_comparisons}

\begin{figure}[t]
    \centering
    \includegraphics[width=0.9\textwidth]{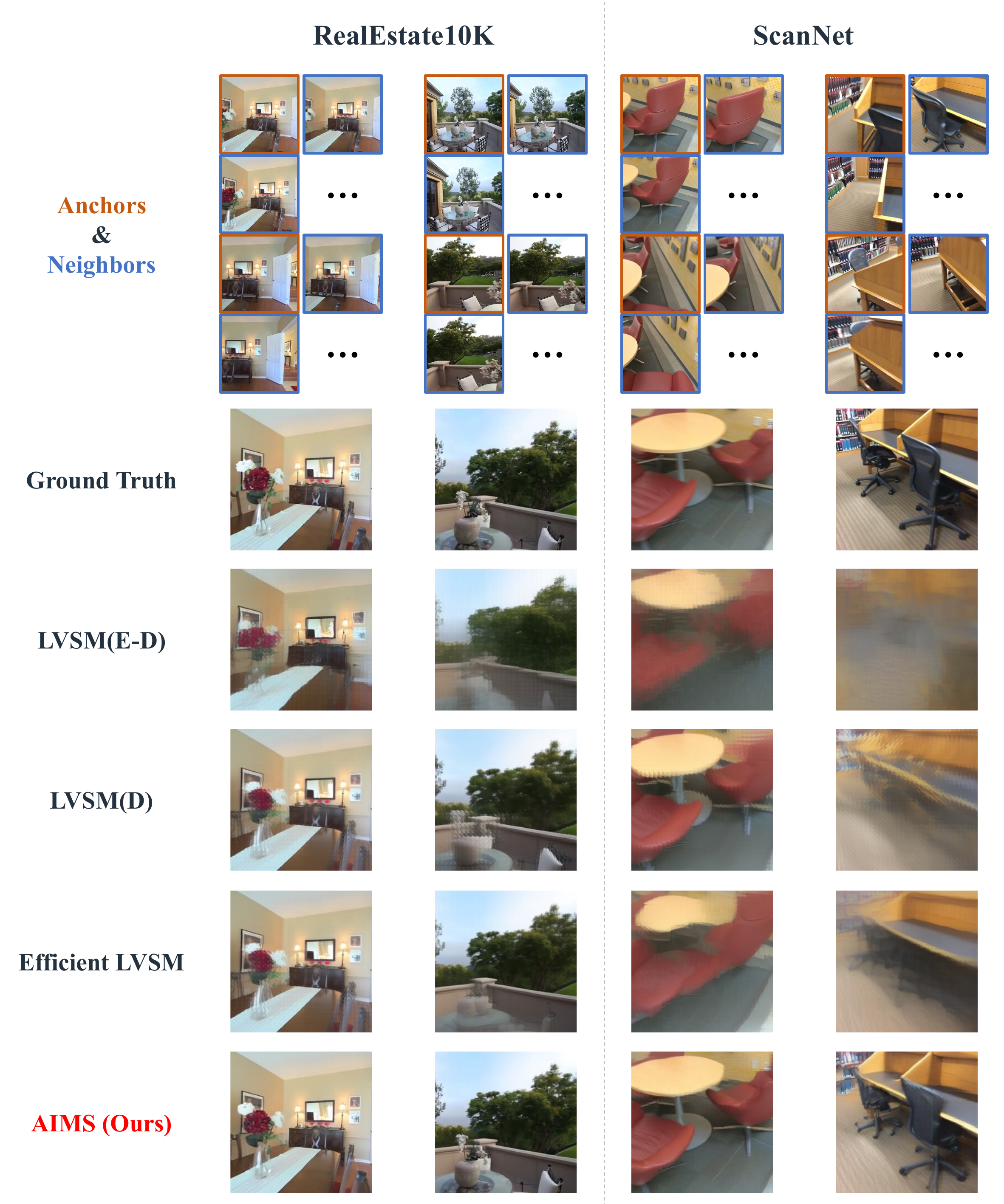} 
    \caption{Qualitative comparison with transformer-based baselines, including LVSM encoder--decoder (E--D), LVSM decoder-only (D), and Efficient-LVSM, on RealEstate10K and ScanNet.}
    \label{fig:qualitative_comparison}
\end{figure}

We compare against the encoder--decoder and decoder-only variants of LVSM~\citep{jin2025lvsm}, Efficient-LVSM~\citep{jia2026efficientlvsm}, MVSplat~\citep{chen2024mvsplat}, and DepthSplat~\citep{xu2025depthsplat} on the RealEstate10K and ScanNet datasets.
For each baseline, we use its officially released RealEstate10K checkpoint.
We likewise train AIMS on RealEstate10K for 20 epochs and evaluate it on both datasets.
See Appendix~\ref{appendix} for additional details on training.
We report PSNR, SSIM, and VGG-based LPIPS~\citep{zhang2018unreasonable}, averaged over target images, alongside inference time and qualitative comparisons (Fig.~\ref{fig:qualitative_comparison}).
Tables~\ref{tab:aims_quality} and~\ref{tab:aims_baseline_anchor_sweep} jointly compare reconstruction quality and inference time using all selected observations and anchor-only baseline inputs, respectively.

\begin{table}[t]
    \centering
    \caption{Reconstruction quality and inference time for baselines evaluated on all observations selected by coverage-aware view grouping with $G=12$ and $K=5$. Encoding is performed once per scene; rendering uses cached representations where available, while LVSM (D) reports full per-view inference. MVSplat fails due to out-of-memory errors. Best results are bold.}
    \label{tab:aims_quality}
    \small
    \setlength{\tabcolsep}{2.5pt}
    \begin{tabular*}{\linewidth}{@{\extracolsep{\fill}}ccccccccc@{}}
        \toprule
        & \multicolumn{3}{c}{RealEstate10K}
        & \multicolumn{3}{c}{ScanNet}
        & Encode & Render \\
        \cmidrule(lr){2-4}\cmidrule(lr){5-7}
        Method & PSNR$\uparrow$ & SSIM$\uparrow$ & LPIPS$\downarrow$
        & PSNR$\uparrow$ & SSIM$\uparrow$ & LPIPS$\downarrow$
        & (ms/scene)$\downarrow$ & (ms/view)$\downarrow$ \\
        \midrule
        DepthSplat
        & 18.28 & 0.6780 & 0.3169
        & 15.70 & 0.5528 & \textbf{0.5118}
        & 1043.04 & 9.84 \\
        MVSplat
        & -- & -- & --
        & -- & -- & --
        & -- & -- \\
        LVSM (E--D)
        & 18.72 & 0.5751 & 0.3896
        & 12.69 & 0.4954 & 0.6916
        & 758.93 & 9.62 \\
        LVSM (D)
        & 24.32 & 0.8009 & 0.1770
        & 17.34 & 0.5698 & 0.5496
        & -- & 1584.35 \\
        Efficient-LVSM
        & 22.06 & 0.7301 & 0.2093
        & 14.68 & 0.5453 & 0.6271
        & 112.08 & 52.25 \\
        \midrule
        \textbf{AIMS (ours)}
        & \textbf{29.41} & \textbf{0.8971} & \textbf{0.0943}
        & \textbf{17.73} & \textbf{0.5774} & 0.5201
        & 161.54 & 7.24 \\
        \bottomrule
    \end{tabular*}
\end{table}

\subsubsection{Reconstruction quality}
\label{sec:aims_quality}

Using all selected observations (Table~\ref{tab:aims_quality}), AIMS achieves $29.41$\,dB PSNR, $0.8971$ SSIM, and $0.0943$ LPIPS on RealEstate10K, outperforming the baselines across all three quality metrics and exceeding decoder-only LVSM by $5.08$\,dB PSNR.
On ScanNet, AIMS achieves the highest PSNR and SSIM in the table, at $17.73$\,dB and $0.5774$, respectively, compared with
$17.34$\,dB PSNR for decoder-only LVSM.
AIMS obtains an LPIPS of $0.5201$, improving on the transformer baselines, while DepthSplat achieves a lower LPIPS of $0.5118$.

\begin{table}[t]
\centering
\caption{Reconstruction quality and inference time for baselines using only $2$, $4$, $8$, or $12$ anchor views. 
AIMS retains $G=12$ anchors and $K=5$ neighbors per anchor. Encoding is performed once per scene; rendering uses cached representations where available, while LVSM (D) reports full per-view inference. Best results are bold.}
\label{tab:aims_baseline_anchor_sweep}
\small
\setlength{\tabcolsep}{2pt}
\begin{tabular*}{\linewidth}{@{\extracolsep{\fill}}cccccccccc@{}}
\toprule
& & \multicolumn{3}{c}{RealEstate10K}
& \multicolumn{3}{c}{ScanNet} & Encode & Render \\
\cmidrule(lr){3-5}\cmidrule(lr){6-8}
Method & Anchors & PSNR$\uparrow$ & SSIM$\uparrow$ & LPIPS$\downarrow$
& PSNR$\uparrow$ & SSIM$\uparrow$ & LPIPS$\downarrow$
& (ms/scene)$\downarrow$ & (ms/view)$\downarrow$ \\
\midrule
\multirow[c]{4}{*}{DepthSplat}
& 2  & 22.97 & 0.8017 & 0.1980 & 11.36 & 0.4345 & 0.6177 & 41.66  & 7.21 \\
& 4  & 24.03 & 0.8598 & 0.1526 & 13.55 & 0.5180 & 0.5754 & 65.15  & 7.30 \\
& 8  & 22.21 & 0.8305 & 0.1834 & 15.21 & 0.5612 & 0.5223 & 114.10 & 8.50 \\
& 12 & 21.19 & 0.8002 & 0.2108 & 15.76 & 0.5757 & \textbf{0.4971} & 172.59 & 9.07 \\
\midrule
\multirow[c]{4}{*}{MVSplat}
& 2  & 21.50 & 0.7590 & 0.2196 & 8.52  & 0.1810 & 0.6437 & 34.08  & 4.11 \\
& 4  & 21.71 & 0.7989 & 0.2084 & 9.94  & 0.2915 & 0.6154 & 58.21  & 4.65 \\
& 8  & 20.21 & 0.7718 & 0.2429 & 11.90 & 0.4087 & 0.5894 & 151.95 & 5.20 \\
& 12 & 19.57 & 0.7551 & 0.2599 & 12.94 & 0.4529 & 0.5818 & 309.39 & 6.16 \\
\midrule
\multirow[c]{4}{*}{LVSM (E--D)}
& 2  & 24.93 & 0.8174 & 0.1818 & 12.81 & 0.5058 & 0.6625 & 10.60 & 9.50 \\
& 4  & 25.60 & 0.8280 & 0.1520 & 13.86 & 0.5239 & 0.6369 & 15.07 & 9.49 \\
& 8  & 22.68 & 0.7387 & 0.2289 & 14.61 & 0.5349 & 0.6247 & 26.25 & 9.50 \\
& 12 & 21.36 & 0.6888 & 0.2760 & 14.71 & 0.5358 & 0.6293 & 42.58 & 9.62 \\
\midrule
\multirow[c]{4}{*}{LVSM (D)}
& 2  & 25.67 & 0.8326 & 0.1629 & 12.78 & 0.5103 & 0.6615 & -- & 12.72 \\
& 4  & 28.02 & 0.8795 & 0.1051 & 14.14 & 0.5291 & 0.6201 & -- & 20.33 \\
& 8  & 28.24 & 0.8896 & 0.0987 & 15.92 & 0.5614 & 0.5691 & -- & 40.07 \\
& 12 & 27.78 & 0.8832 & 0.1039 & 16.85 & \textbf{0.5779} & 0.5473 & -- & 66.43 \\
\midrule
\multirow[c]{4}{*}{Efficient-LVSM}
& 2  & 23.27 & 0.7631 & 0.2012 & 12.72 & 0.5069 & 0.6635 & 7.50 & 8.79 \\
& 4  & 21.43 & 0.7022 & 0.2117 & 13.49 & 0.5200 & 0.6437 & 7.53  & 9.19 \\
& 8  & 21.97 & 0.7239 & 0.2041 & 14.13 & 0.5307 & 0.6303 & 11.61 & 9.01 \\
& 12 & 22.08 & 0.7290 & 0.2054 & 14.38 & 0.5356 & 0.6277 & 14.87 & 10.62 \\
\midrule
\textbf{AIMS (ours)}
& 12 & \textbf{29.41} & \textbf{0.8971} & \textbf{0.0943}
& \textbf{17.73} & 0.5774 & 0.5201 & 161.54 & 7.24 \\
\bottomrule
\end{tabular*}
\end{table}

With anchor-only inputs (Table~\ref{tab:aims_baseline_anchor_sweep}), baselines including encoder--decoder LVSM and DepthSplat achieve their best RealEstate10K results with four anchors.
However, on ScanNet, where camera configurations are more complex, all evaluated baselines achieve higher PSNR as the anchor count increases from two to 12, suggesting that such scenes benefit from broader view coverage.
Yet moving from anchor-only inputs to all selected observations in Table~\ref{tab:aims_quality} yields limited or inconsistent
quality gains for the released baselines, highlighting their difficulty in exploiting larger input collections.

\subsubsection{Computational cost}
\label{sec:aims_cost}

We assess the quality--cost trade-off using the inference times measured on an NVIDIA RTX PRO 6000 Blackwell Server Edition GPU.
AIMS, encoder--decoder LVSM, Efficient-LVSM, MVSplat, and DepthSplat encode the scene once and reuse the resulting representation across target views, whereas decoder-only LVSM performs full inference for each target view.

When all selected observations are used (Table~\ref{tab:aims_quality}), Efficient-LVSM achieves the lowest one-time encoding cost at 112.08\,ms per scene, compared with 161.54\,ms for AIMS. 
Once encoded, however, AIMS achieves the fastest per-view rendering time at 7.24\,ms, corresponding to a $1.33\times$ speedup over the next-fastest method, encoder--decoder LVSM at 9.62\,ms per view.

The anchor-only comparison in Table~\ref{tab:aims_baseline_anchor_sweep} further isolates rendering efficiency under different input budgets. 
Despite integrating substantially more observations, AIMS achieves the fastest per-view rendering time among the transformer-based methods at 7.24\,ms.

These results demonstrate that AIMS effectively integrates larger input sets to deliver strong reconstruction quality while maintaining low per-view rendering cost. 
Its fixed global attention budget offers a favorable trade-off between reconstruction quality and computational cost.

\subsection{Ablation Studies and Analysis}
\label{sec:aims_ablation}

\begin{figure}[t]
    \centering
    \includegraphics[width=\linewidth]{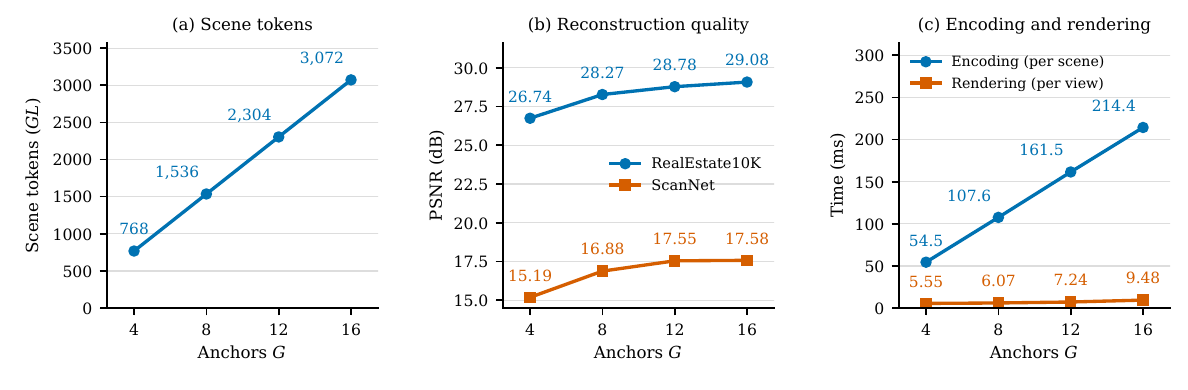}
    \caption{Quality--cost trade-off across anchor counts for controlled 10-epoch training runs with $G\in\{4,8,12,16\}$ and fixed $K=5$. (a) Number of scene tokens, (b) reconstruction quality on RealEstate10K and ScanNet, and (c) scene encoding and per-view rendering time.}
    \label{fig:anchor_ablation}
\end{figure}

\begin{table}[t]
    \centering
    \caption{Effect of inference-time neighborhood size $K$ on reconstruction quality. All settings use the same AIMS model trained with $G=12$ and $K=5$, without retraining. Best results are bold.}
    \label{tab:aims_k_ablation}
    \small
    \setlength{\tabcolsep}{4pt}
    \begin{tabular*}{\linewidth}
    {@{\hspace{8pt}\extracolsep{\fill}}ccccccc@{\hspace{8pt}}}
        \toprule
        & \multicolumn{3}{c}{RealEstate10K}
        & \multicolumn{3}{c}{ScanNet} \\
        \cmidrule(lr){2-4}\cmidrule(lr){5-7}
        $K$ & PSNR$\uparrow$ & SSIM$\uparrow$ & LPIPS$\downarrow$
            & PSNR$\uparrow$ & SSIM$\uparrow$ & LPIPS$\downarrow$ \\
        \midrule
        1 & 29.07 & 0.8916 & 0.0973
          & 16.82 & 0.5649 & 0.5352 \\
        3 & \textbf{29.50} & \textbf{0.8988} & \textbf{0.0932}
          & 17.49 & 0.5749 & 0.5237 \\
        5 & 29.41 & 0.8971 & 0.0943
          & 17.73 & 0.5774 & 0.5201 \\
        7 & 29.23 & 0.8941 & 0.0961
          & \textbf{17.82} & \textbf{0.5779} & \textbf{0.5192} \\
        \bottomrule
    \end{tabular*}
\end{table}

\subsubsection{Effects of anchor count and neighborhood size}
\label{subsubsec:effect_anchor_neighbor}

To examine the trade-off between reconstruction quality and computational cost, we conduct controlled 10-epoch training runs with $G\in\{4,8,12,16\}$ while fixing the neighborhood size to $K=5$.
As shown in Fig.~\ref{fig:anchor_ablation}, reconstruction quality improves as the number of anchors increases, but the gains diminish at larger $G$.
Increasing $G$ from $12$ to $16$ increases the number of scene tokens from $2{,}304$ to $3{,}072$, resulting in higher computational cost: scene encoding time increases from $161.54$ to $214.36$\,ms and per-view rendering time from $7.24$ to $9.48$\,ms.
This corresponds to approximately $33\%$ more scene tokens, $33\%$ higher encoding cost, and $31\%$ higher rendering cost, while PSNR improves by only $0.29$\,dB on RealEstate10K and $0.03$\,dB on ScanNet.
After weighing the performance gains against the additional computational cost, we select $G=12$ as our main configuration.

While $G$ determines the size of the global scene representation, $K$ controls the local context aggregated into each anchor.
We next examine this complementary design choice by varying $K\in\{1,3,5,7\}$ at inference time while keeping $G=12$.
This experiment uses a single 20-epoch checkpoint trained with $G=12$ and $K=5$---the same checkpoint used in our main baseline comparison---without retraining.
As shown in Table~\ref{tab:aims_k_ablation}, the effect of neighborhood size differs across datasets.
On RealEstate10K, which predominantly contains smooth camera trajectories and relatively constrained viewpoint variation, reconstruction quality peaks at $K=3$, with larger neighborhoods providing no further improvement.
In contrast, ScanNet exhibits consistent gains as $K$ increases from $1$ to $7$: PSNR increases from $16.82$ to $17.82$\,dB, while SSIM improves from $0.5649$ to $0.5779$ and LPIPS decreases from $0.5352$ to $0.5192$.
This suggests that larger local neighborhoods provide useful complementary observations in more complex scenes, where individual anchors may capture insufficient local scene information.

\begin{figure}[t]
    \centering
    \includegraphics[width=\linewidth]{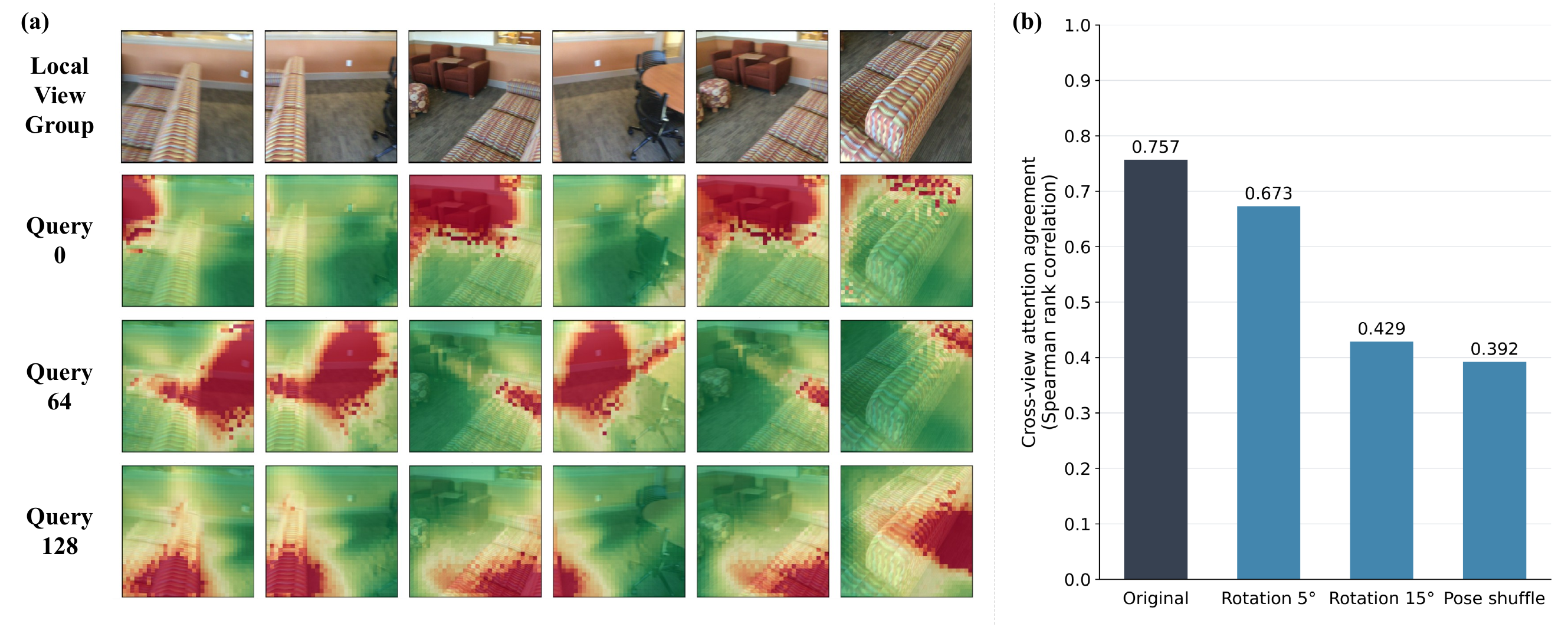}
    \caption{Analysis of VGE attention and its sensitivity to camera pose. (a) Head-averaged attention maps at VGE layer 6 for learnable queries 0, 64, and 128 across a local view group. (b) Cross-view attention agreement under camera-pose perturbations, measured by the mean Spearman rank correlation over matched physical locations across 50 ScanNet scenes.}
    \label{fig:cam_geo_influence}
\end{figure}

\subsubsection{Analyzing camera-pose utilization in VGE attention}
\label{subsubsec:attention_analysis}

VGE aggregates information from multiple observations within each view group.
To understand what guides this aggregation, we begin by visualizing how its learnable queries attend to the input views.
Fig.~\ref{fig:cam_geo_influence}(a) shows the head-averaged attention maps of queries 0, 64, and 128 at VGE layer 6.
The queries exhibit distinct spatial preferences, emphasizing the armchairs, the wall near the electrical outlet, and the
sofa corner, respectively.
These preferences recur across views in which the corresponding regions are visible.
We therefore ask whether these patterns reflect image appearance alone or whether camera pose also guides cross-view aggregation.

To investigate this, we randomly select 50 of the 843 ScanNet scenes and construct 12 anchor--neighbor groups per scene, each containing one anchor and five neighbors, yielding 600 groups.
Within each group, we use depth and camera poses to identify patches observing the same scene locations.
For each learnable query at layer 6, we compute Spearman's rank correlation~\citep{spearman1987proof} between attention values at these matched patches, measuring how consistently the query prioritizes the same locations across views.
We then repeat the analysis while modifying only the camera poses supplied to the model.
Each camera orientation is rotated by $5^\circ$ or $15^\circ$ about an independently sampled random axis while keeping its position fixed.
We also shuffle complete poses among views.
The images, view groups, and patch correspondences established from the original poses remain fixed across all conditions.

As shown in Fig.~\ref{fig:cam_geo_influence}(b), the mean correlation decreases from $0.757$ with the original poses to $0.673$ and $0.429$ under $5^\circ$ and $15^\circ$ rotations, respectively, and to $0.392$ under pose shuffling.
This decline despite unchanged image content suggests that camera pose plays a nontrivial role in shaping VGE's cross-view aggregation.
We hypothesize that this sensitivity enables VGE to encode aggregated scene information in a more spatially aware manner, yielding more informative anchor representations and contributing to AIMS's reconstruction performance.
Further details on the analysis procedure are provided in Appendix~\ref{app:attention_computation}.

\section{Conclusion}
\label{sec:conclusion}

In this work, we present AIMS, a transformer-based novel view synthesis framework that addresses the growing attention cost of
large input collections. By combining coverage-aware view grouping with the View Group Encoder, AIMS integrates anchors and neighboring observations into compact scene tokens for synthesis under a fixed global attention budget.
Experiments on RealEstate10K and ScanNet demonstrate high-quality reconstructions with low per-view rendering cost.
These results establish hierarchical view integration as an effective approach to balancing reconstruction quality and computational cost in scalable novel view synthesis.




\bibliography{iclr2027_conference}
\bibliographystyle{iclr2027_conference}

\appendix
\section{Appendix}
\label{appendix}

\subsection{AIMS Implementation and Training Details}
\label{app:aims_implementation}

Images are resized and center-cropped to $256\times256$, with camera intrinsics adjusted accordingly and poses normalized to
a common scene frame.
We use $8\times8$ image patches, as in LVSM~\citep{jin2025lvsm} and Efficient-LVSM~\citep{jia2026efficientlvsm}.
The View Group Encoder (VGE) and global decoder each contain 12 transformer blocks with hidden dimension $D=768$ and
12 attention heads.
Each group produces $L=192$ scene tokens.
Our main configuration uses $G=12$ anchors and $K=5$ neighbors per anchor, yielding six-view groups and $GL=2,304$ scene tokens.
Both anchor selection and neighborhood construction use pose-distance weights $w_{\mathrm{t}}=1$ and $w_{\mathrm{r}}=0.25$.

We initialize the source image tokenizer and decoding modules, including the target ray tokenizer and RGB prediction head,
from LVSM's pretrained encoder--decoder checkpoint. 
The source image tokenizer remains frozen, while the randomly initialized VGE and pretrained decoding modules are optimized jointly. 
Training uses AdamW with learning rates of $2\times10^{-4}$ for the VGE and $5\times10^{-5}$ for the decoding modules,
$(\beta_1,\beta_2)=(0.9,0.95)$, and weight decay $0.05$. 
We train for 20 epochs with 500 linear warm-up steps followed by cosine decay to $5\%$ of the initial learning rates.
Training uses five NVIDIA RTX PRO 6000 Blackwell Server Edition GPUs, with 20 scenes per GPU and a global batch size of 100,
without gradient accumulation.
We use BF16 mixed precision, gradient checkpointing, gradient clipping at a norm of $1$, and random seed 42.
The training objective combines MSE and VGG-based LPIPS with perceptual-loss weight $\lambda=0.5$.

\begin{figure}[t]
    \centering
    \includegraphics[width=\linewidth]{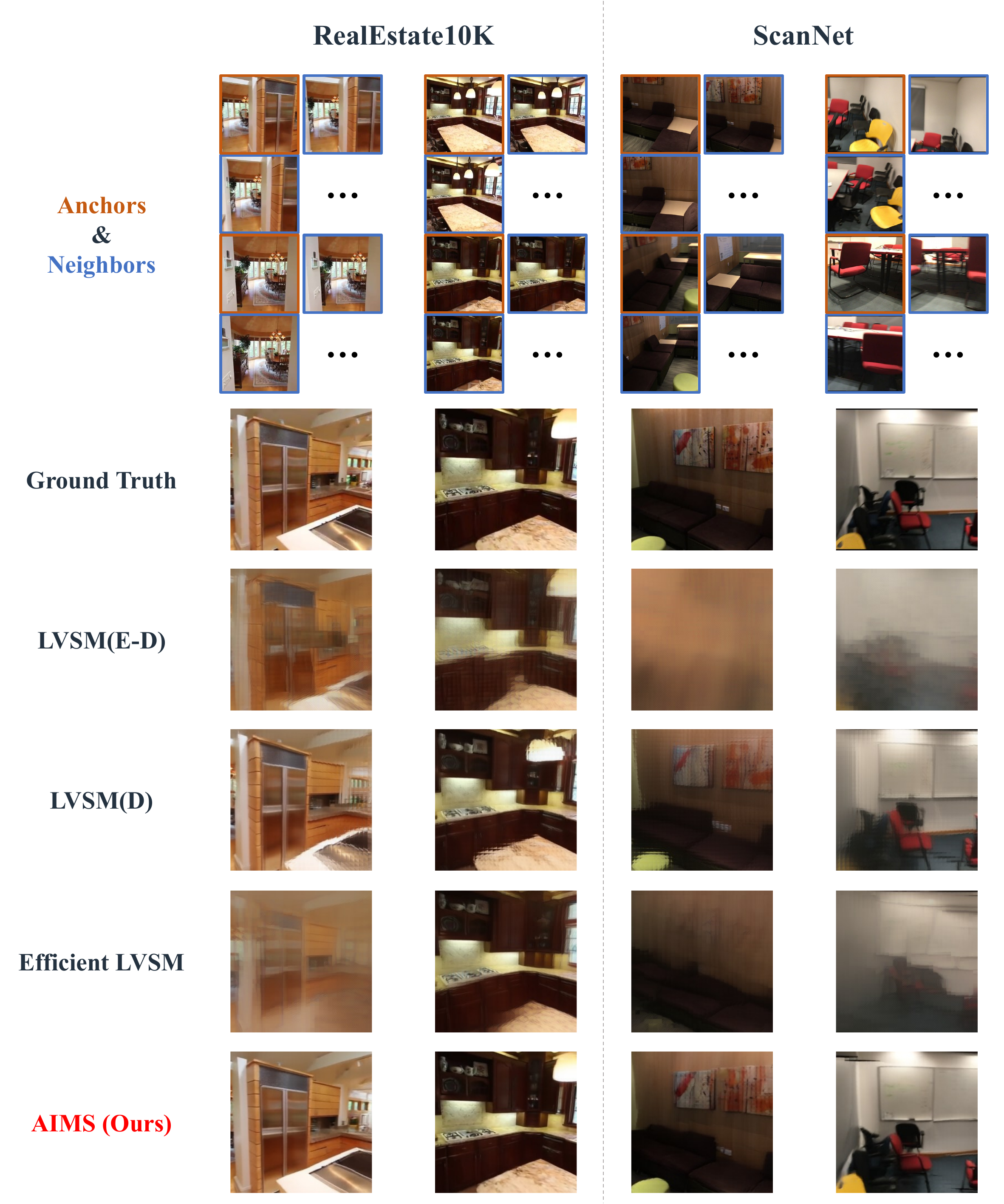}
    \caption{Additional qualitative comparisons with transformer-based baselines, including LVSM encoder--decoder (E--D), LVSM decoder-only (D), and Efficient-LVSM, on RealEstate10K and ScanNet.}
    \label{appfig:additional_view_synthesis}
\end{figure}

We train on 66,033 RealEstate10K clips.
For each scene, we uniformly sample 100 views to form the candidate view set.
During training, we sample up to eight target views at random from this set, while validation uses up to eight uniformly spaced target views.
In contrast to the evaluation setting in the main paper, we permit source--target overlap during training.
We select the checkpoint with the highest validation PSNR, which occurs at epoch 20.

For the anchor-count study, each configuration is trained for 10 epochs with $K=5$.
Gradient accumulation uses four steps for $G=4,8$ and two steps for $G=12,16$.

\subsection{Implementation Details of the Attention Analysis}
\label{app:attention_computation}

\paragraph{Setup and attention extraction.}
We use the same trained checkpoint as in the main comparison.
The fixed evaluation cohort contains 50 distinct ScanNet scenes.
For each scan, we load 100 uniformly sampled frames and exclude eight held-out target frames before applying the standard
AIMS grouping procedure.
We evaluate all 12 groups per scene, each containing one anchor and five neighbors.
Images are resized and cropped to $256\times256$, producing $32\times32$ grids of $8\times8$ patches.
We extract attention from all 192 learnable queries at the sixth VGE layer and average over attention heads.
We use the original softmax probabilities, without separately normalizing the attention map of each image.

\paragraph{Depth-based patch matching.}
Depth is used only to establish correspondences, not as an input to VGE.
We retain depths between $0.1$ and $10$\,m and require at least $25\%$ valid depth pixels within a patch.
To avoid patches spanning large depth discontinuities, we reject those whose 90th--10th percentile depth difference exceeds
$0.10 + 0.05d$\,m, where $d$ is the patch's median depth in meters.
For each remaining patch, we select a valid, median-consistent depth sample near its center, back-project it into 3D, and project it into another view using the original camera poses.
A match must fall inside that image and satisfy
\[
    |z_{\mathrm{projected}}-z_{\mathrm{observed}}|
    \leq 0.05\,\mathrm{m}+0.02z_{\mathrm{projected}}.
\]
We retain at most one source-patch match per destination patch.
Unmatched or occluded regions are excluded rather than assigned zero attention.

\paragraph{Pose perturbations.}
For the rotation conditions, we independently sample a random axis in each camera's local coordinate system and rotate its
orientation by $5^\circ$ or $15^\circ$, keeping its position fixed.
The two rotation magnitudes use the same sampled axes. 
For pose shuffling, we reassign complete poses among the group's unique views so that no image retains its original pose.
Perturbations are applied after the standard pose normalization, without subsequent renormalization.
Images, intrinsics, group membership, and reference patch matches remain unchanged across conditions.

\paragraph{Correlation and averaging.}
For each query and ordered pair of views, we compute Spearman's correlation between the attention values at matched patches,
using average ranks for ties.
We require at least 32 valid matches and exclude nearly constant attention vectors, for which the correlation is unreliable.
Queries assigning less than $10^{-4}$ total attention to source-image patches are also excluded.
Valid correlations are averaged first over view pairs for each query, then over queries within each group, and finally over
groups within each scene.
The reported score is the arithmetic mean of the resulting 50 scene scores, giving each scene equal weight.

\begin{figure}[t]
    \centering
    \includegraphics[width=\linewidth]{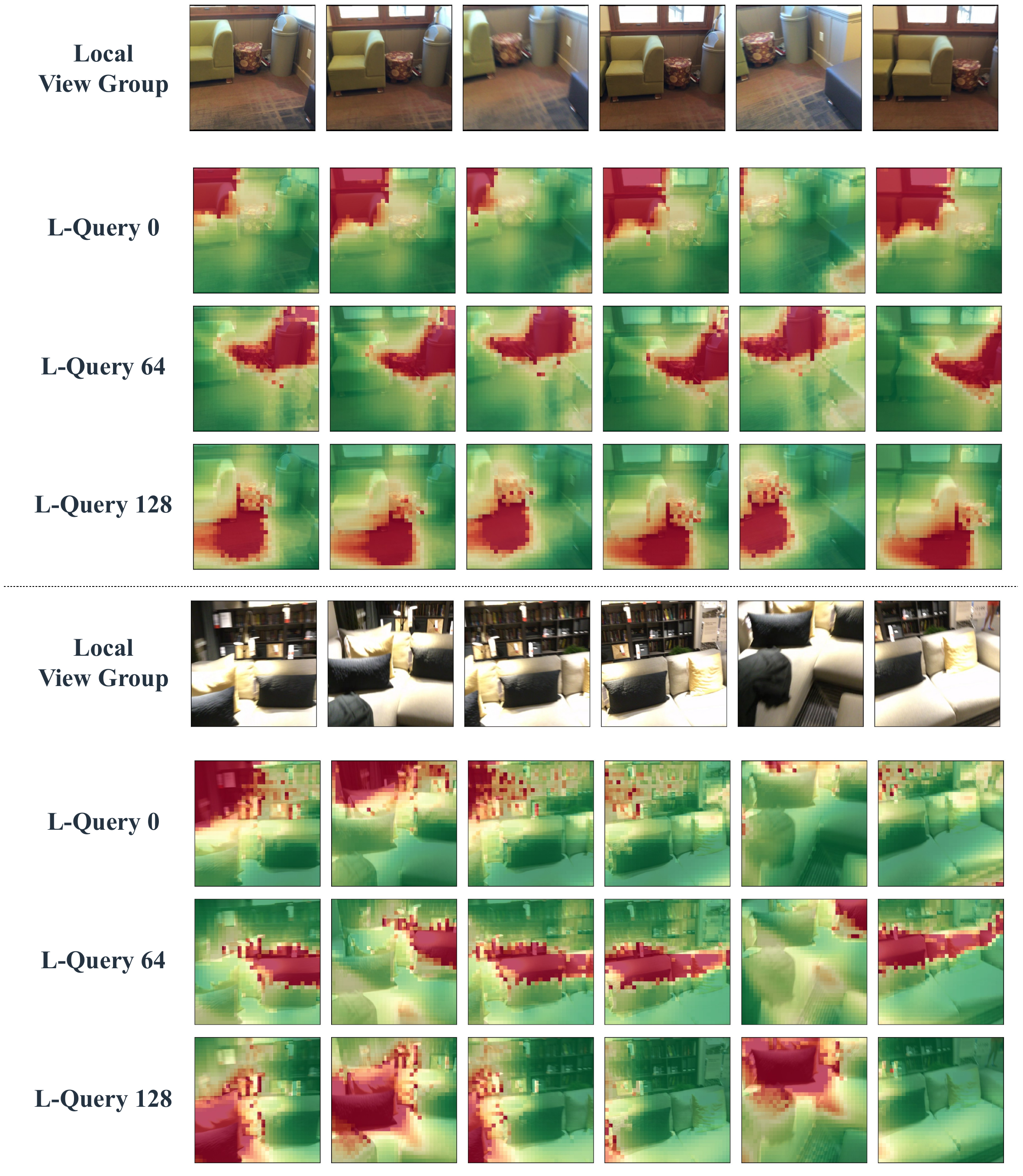}
    \caption{More visualizations on head-averaged attention maps at VGE layer 6 for learnable queries 0, 64, and 128 across a local view group.}
    \label{appfig:additional_attention_maps}
\end{figure}

\subsection{Additional View Synthesis Results}
Fig.~\ref{appfig:additional_view_synthesis} presents additional novel view synthesis results comparing AIMS with transformer-based baselines.

\subsection{Additional attention map visualizations}
Fig.~\ref{appfig:additional_attention_maps} presents more visualizations on VGE attention maps for the given local view group.

\subsection{Limitations and Future Work}
\label{app:aims_limitations}

AIMS maintains a fixed global decoding budget, but view grouping and local integration incur additional computation as the number of views increases.
Moreover, compressing each group into a fixed number of tokens may discard fine detail in complex scenes.
Future work could improve encoding efficiency and allocate scene tokens adaptively according to scene complexity.

Our grouping strategy relies on calibrated camera poses and does not explicitly account for visibility or image content.
Nearby cameras may provide redundant observations, while fixed-size neighborhoods do not guarantee that every available
view contributes to the scene representation.
Combining pose coverage with appearance or visibility cues could improve view selection and integration.

\end{document}